\documentclass[11pt]{article}
\pdfoutput=1
\usepackage{acl2016}
\usepackage{times}
\usepackage{latexsym}
\usepackage{graphicx}

\usepackage{amsmath}
\usepackage{amsfonts}
\usepackage{times}
\usepackage{url}
\usepackage{latexsym}
\usepackage{graphicx}
\usepackage{verbatim}
\usepackage{subfig}
\usepackage{qtree}
\usepackage{graphicx}
\usepackage{caption}
\usepackage{multirow}
\usepackage{mathrsfs}
\usepackage{tablefootnote}
\usepackage{arydshln}
\usepackage{lscape}
\usepackage{afterpage}
\usepackage{todonotes}

\usepackage{pgfplots}
\usetikzlibrary{patterns}
\pgfplotsset{compat=1.5}

\aclfinalcopy 

\title{Quantifying the vanishing gradient and long distance dependency problem in recursive neural networks and recursive LSTMs}

\author{Phong Le \and Willem Zuidema\\
Institute for Logic, Language and Computation \\
University of Amsterdam, the Netherlands\\
{\tt \{p.le,zuidema\}@uva.nl } \\ }

\date{}

\begin{document}

\maketitle

\begin{abstract}
Recursive neural networks (RNN) and their recently proposed extension recursive 
long short term memory networks (RLSTM) are models that compute representations 
for sentences, by recursively combining word embeddings according to an externally 
provided parse tree. Both models thus, unlike recurrent networks, explicitly make 
use of the hierarchical structure of a sentence. In this paper, we demonstrate
that RNNs nevertheless suffer from the vanishing gradient and long distance 
dependency problem, and that RLSTMs greatly improve over RNN's on these problems. 
We present an artificial learning task that allows us to quantify the severity of 
these problems for both models. We further show that a ratio of gradients (at the 
root node and a focal leaf node) is highly indicative of the success of backpropagation
at optimizing the relevant weights low in the tree. This paper thus provides an 
explanation for existing, superior results of RLSTMs on tasks such as sentiment 
analysis, and suggests that the benefits of including hierarchical structure and of 
including LSTM-style gating are complementary.
%
\end{abstract}

\section{Introduction}
\label{sec intro}

The recursive neural network (RNN) model became popular 
since the work of 
\newcite{socher_learning_2010}. 
It has been employed to tackle several NLP tasks, 
such as syntactic parsing \cite{socher2013parsing}, machine translation
\cite{liu-EtAl:2014:P14-1}, and word embedding learning \cite{luong2013better}.
However, like traditional \emph{recurrent} neural networks, the RNN seems  
to suffer from the vanishing gradient problem, in which error signals propagating 
from the root in a parse tree to the child nodes shrink very quickly. Moreover, 
it encounters difficulties in capturing long range dependencies: 
information propagating from child nodes deep in a parse tree can be obscured 
before reaching the root node.

In the recurrent neural network world, the long short term memory (LSTM)
architecture \cite{hochreiter1997long} is often used as a solution to 
these two problems. A natural extension of the LSTM can be defined for tree structures,  
which we call Recursive LSTM (RLSTM), as proposed independently by
\newcite{tai2015improved}, \newcite{zhu2015long}, 
and \newcite{le2015composition}. 
However, while there is intensive research showing how 
the LSTM 
architecture can overcome those two problems compared to traditional 
recurrent models
(e.g., \newcite{gers2001lstm}), 
such research is, to our knowledge, still absent for the comparison between RNNs and RLSTMs. 
Therefore, in the current paper we investigate
the following two questions:
\begin{enumerate}
\item[1.] Is the RLSTM more capable of capturing long range dependencies 
    than the RNN?
\item[2.] Does the RLSTM overcome the vanishing gradient problem 
    more effectively than the RNN?
\end{enumerate}

Supervised learning requires annotated data, which is often 
expensive to collect. As a result, 
examining a model on natural data on many different aspects can be difficult 
because the portion of data that fits a specific aspect 
could not be sufficient. Moreover, studying
individual aspects separately is hard since many aspects are often correlated with each other.
This, unfortunately, is true in our case:
answering those two questions requires us to evaluate the examined models on 
datasets of different tree depths, in which the key nodes which contain 
decisive information in a parse tree must 
be identified. Using available annotated corpora such as the 
Stanford Sentiment Treebank \cite{socher2013recursive} and the Penn Treebank
is thus inappropriate, as they are too small for this purpose (10k, 40k trees, respectively,
compared to 240k trees in our experiments), and key nodes are not marked. 
Our solution is an artificial task where sentences and parse 
trees can be randomly generated under any arbitrary constraints on 
tree depth and key node's position. 

\section{Background}

Both the RNN and the RLSTM model are instances of a general framework which 
takes a sentence, syntactic tree, 
and vector representations for the words in the sentence as input, 
and applies a composition function to recursively compute 
vector representations for all the phrases in the tree and the complete sentence.
Technically speaking, given a production $p \rightarrow x \; y$, 
and $\mathbf{x}, \mathbf{y} \in \mathbb{R}^n$ representing $x,y$, 
we compute $\mathbf{p} \in \mathbb{R}^n$ for $p$ by  
$\mathbf{p} = F(\mathbf{x},\mathbf{y})$,
where $F$ is a composition function.

In the RNN, $F$ is a one-layer feed-forward neural network. 
In the RLSTM, a node $u$ is represented by the vector 
$\mathbf{u} = [\mathbf{u}_{r};\mathbf{u}_{c}]$ resulting from 
concatenating a vector representing the phrase that the node covers 
and a memory vector. $F$ could be any LSTM that can combine two such 
concatenation vectors, such as Structure-LSTM \cite{zhu2015long}, 
Tree-LSTM \cite{tai2015improved}, and LSTM-RNN \cite{le2015composition}.
In the current paper, we use the
implementation\footnote{\url{https://github.com/lephong/lstm-rnn}}
of \newcite{le2015composition}.


\section{Experiments}

\begin{figure*}
\centering
\includegraphics[width=0.8\textwidth]{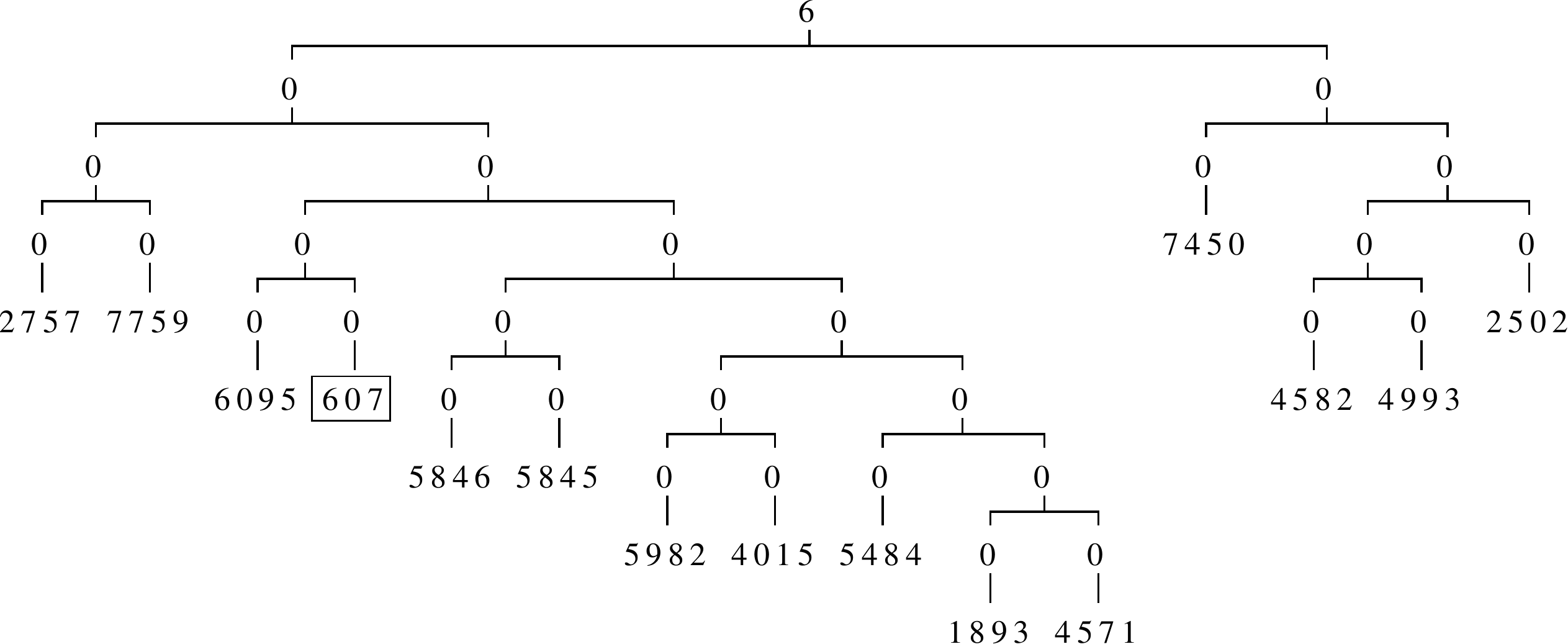}
\caption{Example binary tree for the artificial task. 
The number enclosed in the box is the \emph{keyword} of the sentence.}
\label{fig lstm.faketree}
\end{figure*}

We now examine how the two problems, the vanishing gradient problem and 
the problem of how to capture long range dependencies, affect the  
RLSTM model and the RNN model. 
To do so, we propose the following artificial task, which requires a model
to distinguish useful signals from noise.
We define:
\begin{itemize}
    \item a sentence is a sequence of tokens which are integer numbers 
in the range $[0,10000]$;
    \item a sentence contains one and only one \emph{keyword} token which is 
an integer number smaller than 1000;
    \item a sentence is labeled with the integer resulting from dividing the keyword by 100.
    For instance, 
if the keyword is 607, the label is 6. In this way, there are 10 classes, ranging from 0 to 9.
\end{itemize}
The task is to predict the class of a sentence, given its binary parse tree 
(Figure~\ref{fig lstm.faketree}).
Because the label of a sentence is determined solely by the keyword, the two models need to 
identify the keyword in the parse tree and allow only the information from the leaf node 
of the keyword to affect the root node. 
It is worth noting that this task resembles sentiment analysis with simple cases 
in which the sentiment of a whole sentence is determined 
by one keyword (e.g. ``I like the movie''). 
Simulating complex cases involving negation, composition, etc.  
is straightforward and for future work. But here we believe that the current task 
is adequate to answer our two questions raised in Section~\ref{sec intro}.

The two models, RLSTM and RNN, were implemented with the dimension of vector representations 
and vector memories 50. Following \newcite{socher2013recursive}, 
we used $tanh$ as the activation function, 
and initialized word vectors by randomly sampling each value 
from a uniform distribution $U(-0.0001,0.0001)$. 
We trained the two models using the AdaGrad method
\cite{duchi2011adaptive} with a learning rate of 0.05 and a mini-batch size of 20 for the RNN 
and of 5 for the RLSTM. Development sets were employed for early stopping 
(training is halted when the accuracy on the development set 
is not improved after 5 consecutive epochs).


\subsection{Experiment 1}

We randomly generated 10 datasets. To generate a sentence of length $l$, 
we shuffle a list of randomly chosen $l-1$ non-keywords and one keyword.
The $i$-th dataset contains 12k sentences of lengths
from $10i-9$ tokens to $10i$ tokens, and is split into train, dev, test sets 
with sizes of 10k, 1k, 1k sentences. 
We parsed each sentence by randomly generating a binary tree 
whose number of leaf nodes equals to the sentence length. 

\begin{figure*}[!ht]
\centering
\includegraphics[width=0.6\textwidth]{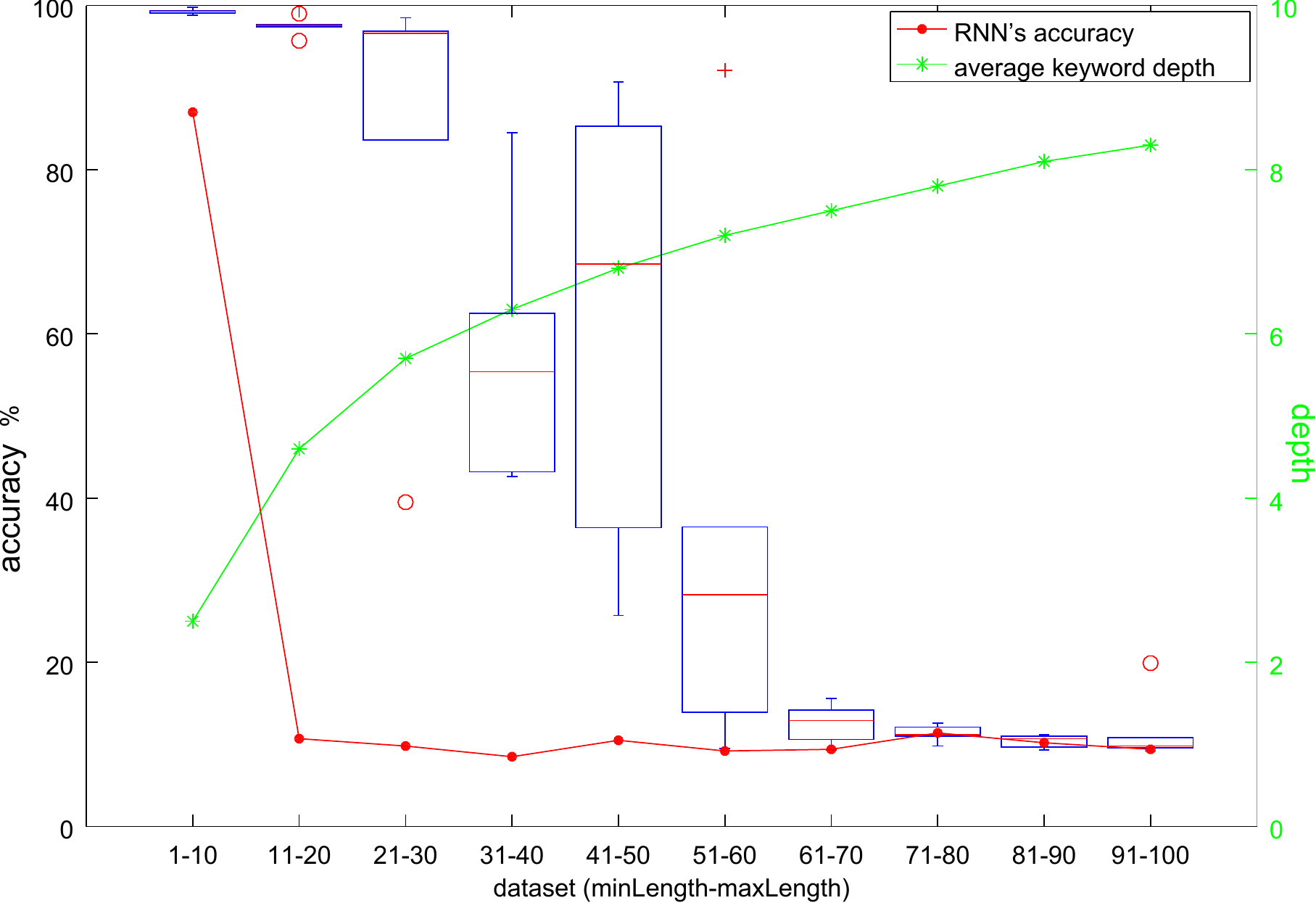}
\caption{Test accuracies of the RNN (the best among 5 runs) and the RLSTM 
(boxplots) on datasets of different sentence lengths.}
\label{fig lstm.rnn fakedata acc}
\end{figure*}

The test accuracies of the two models on the 10 datasets are shown 
in Figure~\ref{fig lstm.rnn fakedata acc}; 
For each dataset we run each model 5 times and reported the highest accuracy for the RNN model,
and the distribution of accuracies (via boxplot) for the RLSTM model.
We can see that the RNN model performs reasonably well on very short sentences 
(less than 11 tokens). However, 
when the sentence length exceeds 10, the RNN's performance drops so quickly that 
the difference between it and the random guess' performance (10\%) is negligible. 
Trying different learning rates, mini-batch sizes, and values for $n$ (the dimension of vectors) 
did not give 
significant differences. On the other hand, the RLSTM model achieves more than 
90\% accuracy on
sentences shorter than 31 tokens. Its performance drops when the sentence length increases, 
but is still substantially better than the random guess when the sentence length 
does not exceed 70. When the sentence length exceeds 70, both the RLSTM and RNN perform
similarly.


\subsection{Experiment 2}

In Experiment 1, it is not clear whether the tree size or the keyword depth is the main 
factor of the rapid drop of the RNN's performance. In this experiment, we kept the tree size 
fixed and vary the keyword depth. 
We generated a pool of sentences of lengths from 21 to 30 tokens
and parsed them by randomly generating binary trees. We then created 10 datasets each of
which has 12k trees (10k for training, 1k for development, and 1k for testing). 
The $i$-th dataset consists of only trees in which distances from
keywords to roots are $i$ or $i+1$ 
(to stop the networks from exploiting keyword depths directly).

\begin{figure*}[!ht]
\centering
\includegraphics[width=0.6\textwidth]{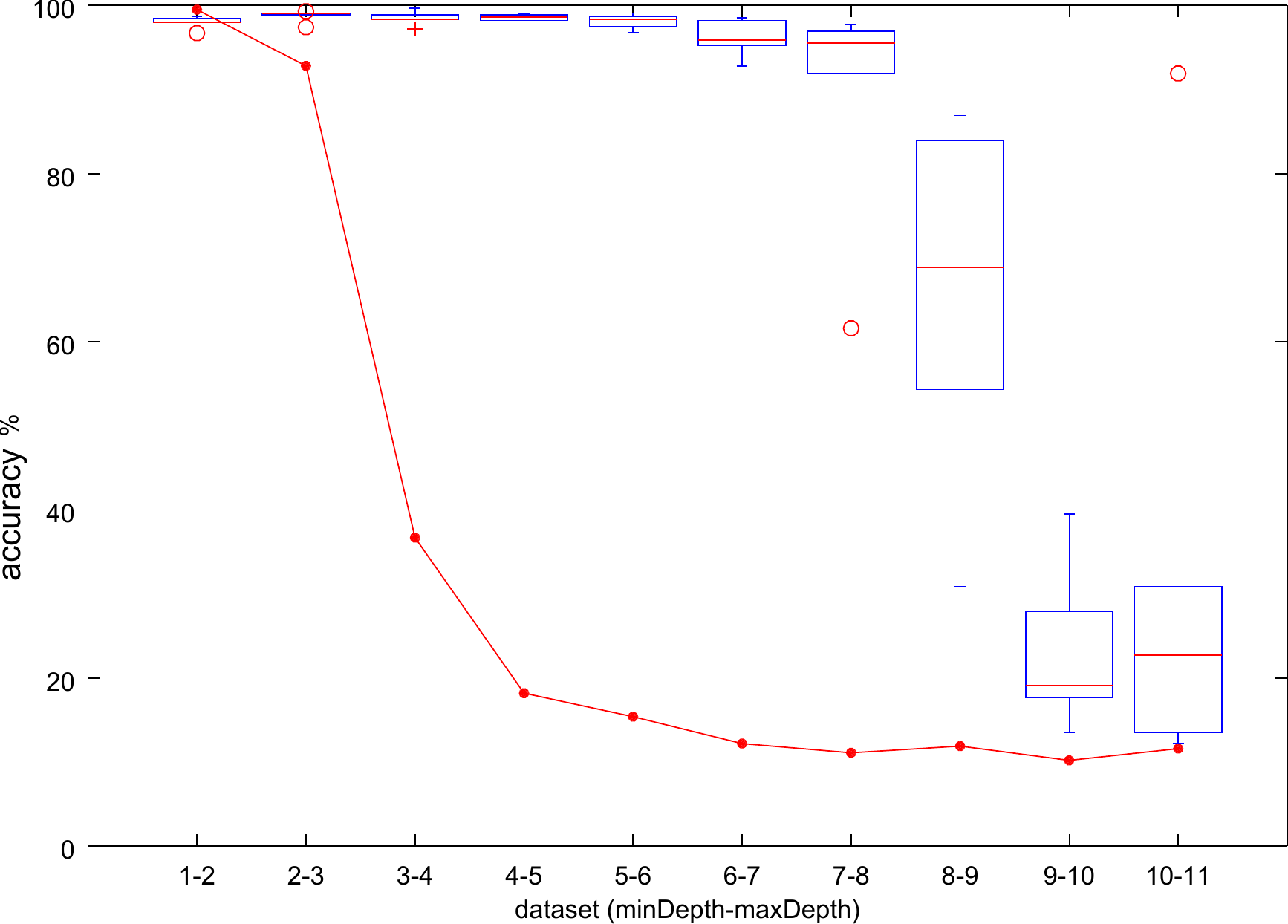}
\caption{Test accuracies of the RNN (the best among 5 runs) 
and the RLSTM (boxplots) on datasets of different keyword depths.}
\label{fig lstm.rnn keywordDepth acc}
\end{figure*}

Figure~\ref{fig lstm.rnn keywordDepth acc} shows test accuracies of the two models on those 10
datasets. Similarly in Experiment 1, for each dataset we run each model 5 times and 
reported the highest accuracy for the RNN model, 
and the distribution of accuracies for the RLSTM model. 
As we can see, the RNN model achieves very high accuracies when the keyword depth 
does not exceed 3. Its performance then drops rapidly and gets close to the
performance of the random guess. This is evidence that the RNN model has difficulty 
capturing long range dependencies. By contrast, the RLSTM model performs at above 90\% accuracy 
until the depth of the keyword reaches 8. 
It has difficulty dealing with larger depths, but the performance 
is always better than the random guess.


\subsection{Experiment 3}

\begin{figure*}
\centering
\includegraphics[width=0.7\textwidth]{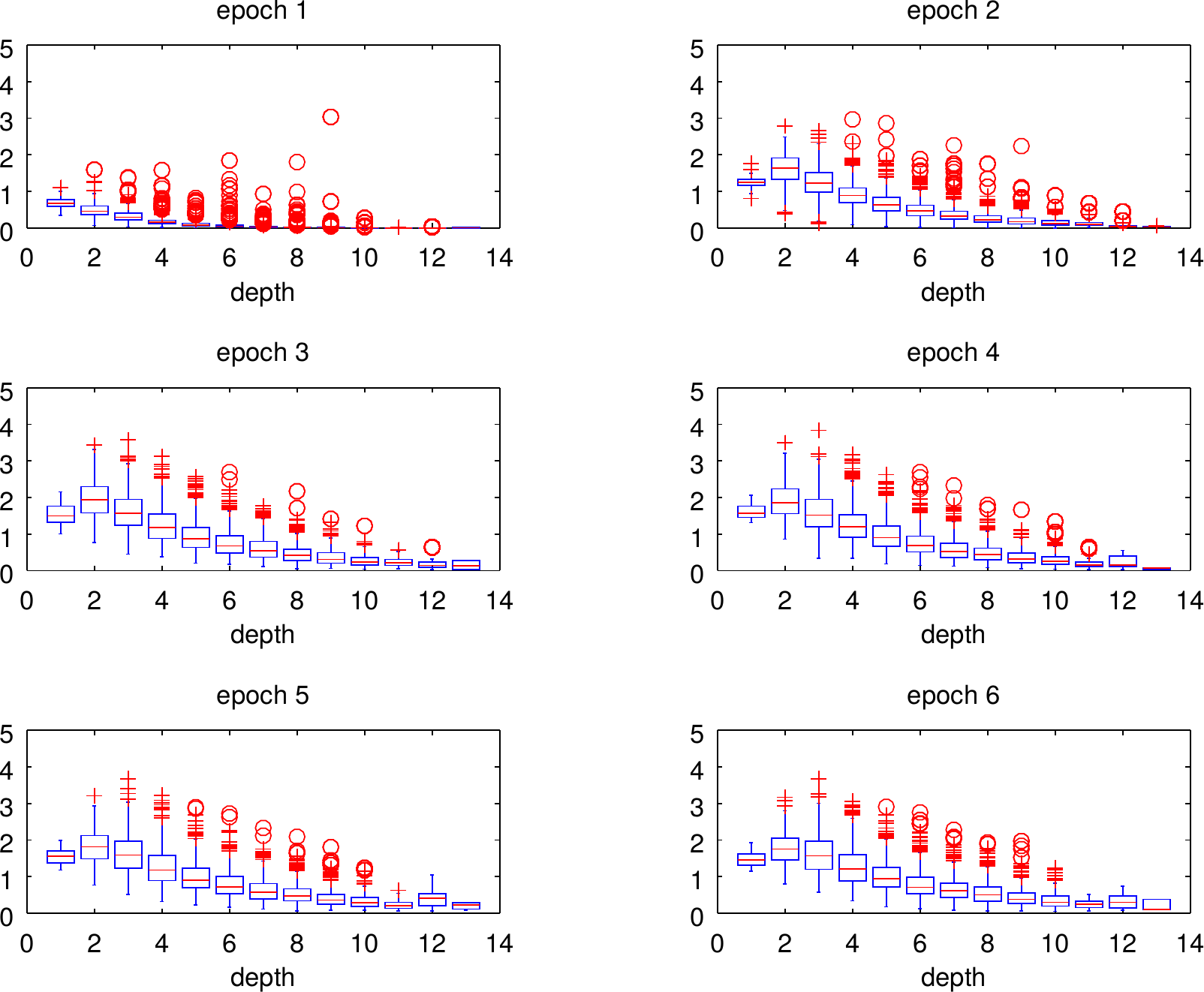}
\caption{Ratios of norms of error vectors at keyword nodes to norms of error vectors 
at root nodes w.r.t. the keyword node depth in each epoch of training the RNN. Gradients gradually vanish with greater depth.}
\label{fig lstm.rnn backprop}
\end{figure*}

\begin{figure*}
\centering
\includegraphics[width=0.7\textwidth]{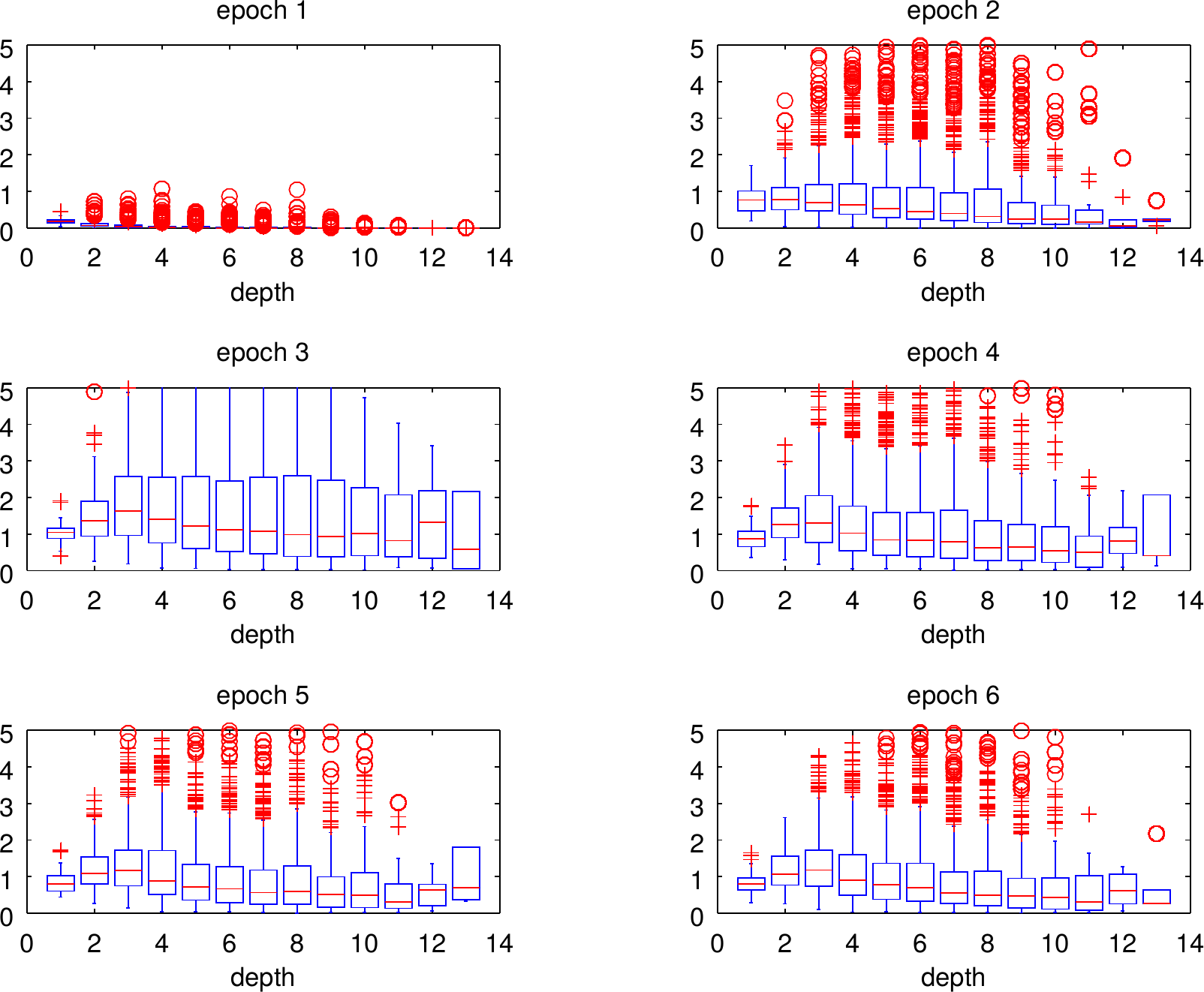}
\caption{Ratios of norms of error vectors at keyword nodes (at different depths) to norms of error vectors 
at root nodes, in the RLSTM. Many gradients explode in epoch 2, but stabilize later. Gradients do not vanish, even at depth 12 and 13. }
\label{fig lstm.lstm backprop}
\end{figure*}

\begin{figure*}
        \centering
        \subfloat[RNN]{%
        \hspace{-.5cm}
                \includegraphics[width=0.6\textwidth]{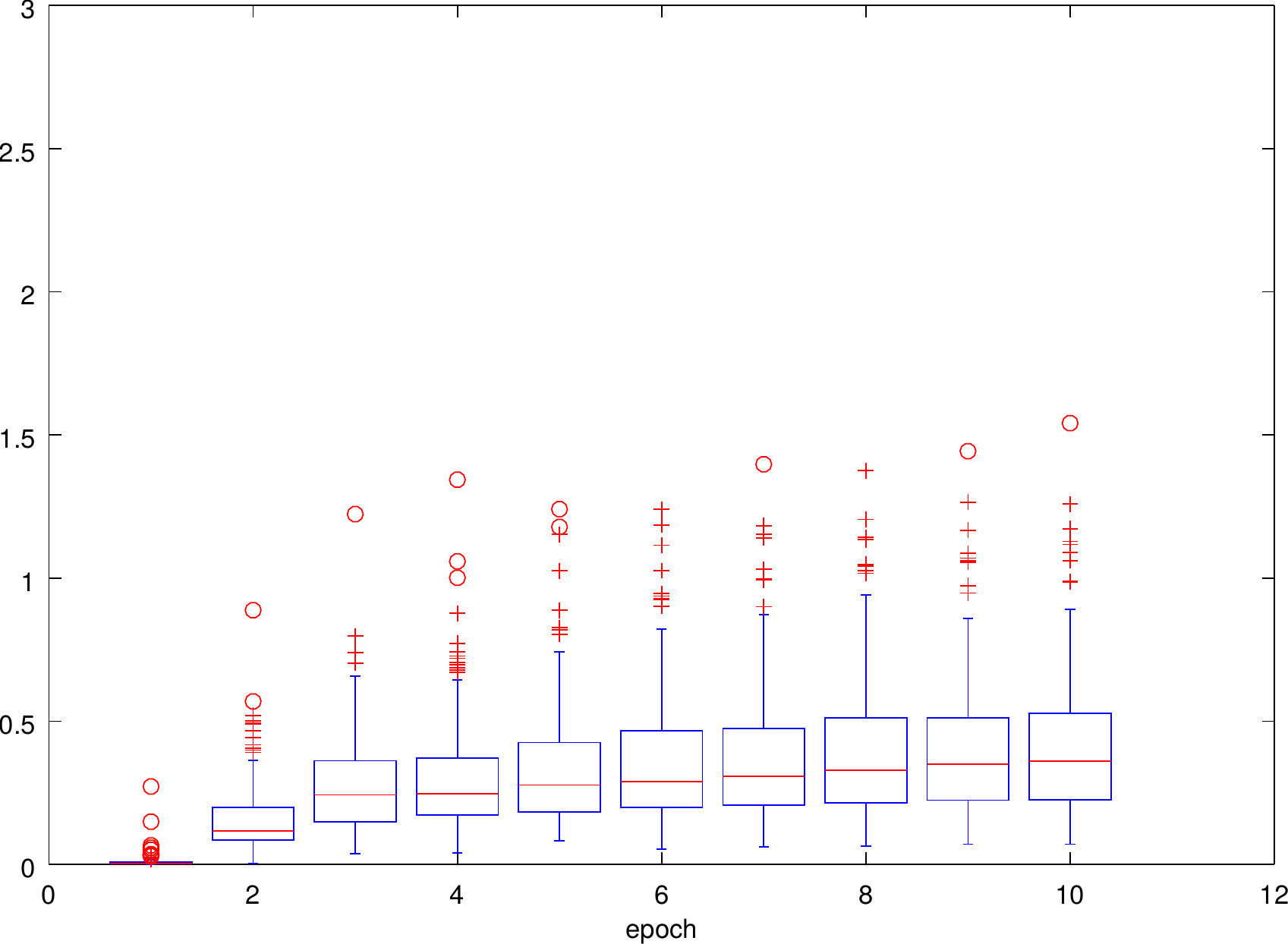}
                \label{fig lstm.rnn depth10}
        }
        
        \subfloat[RLSTM (with development accuracies)]{%
                \includegraphics[width=0.6\textwidth]{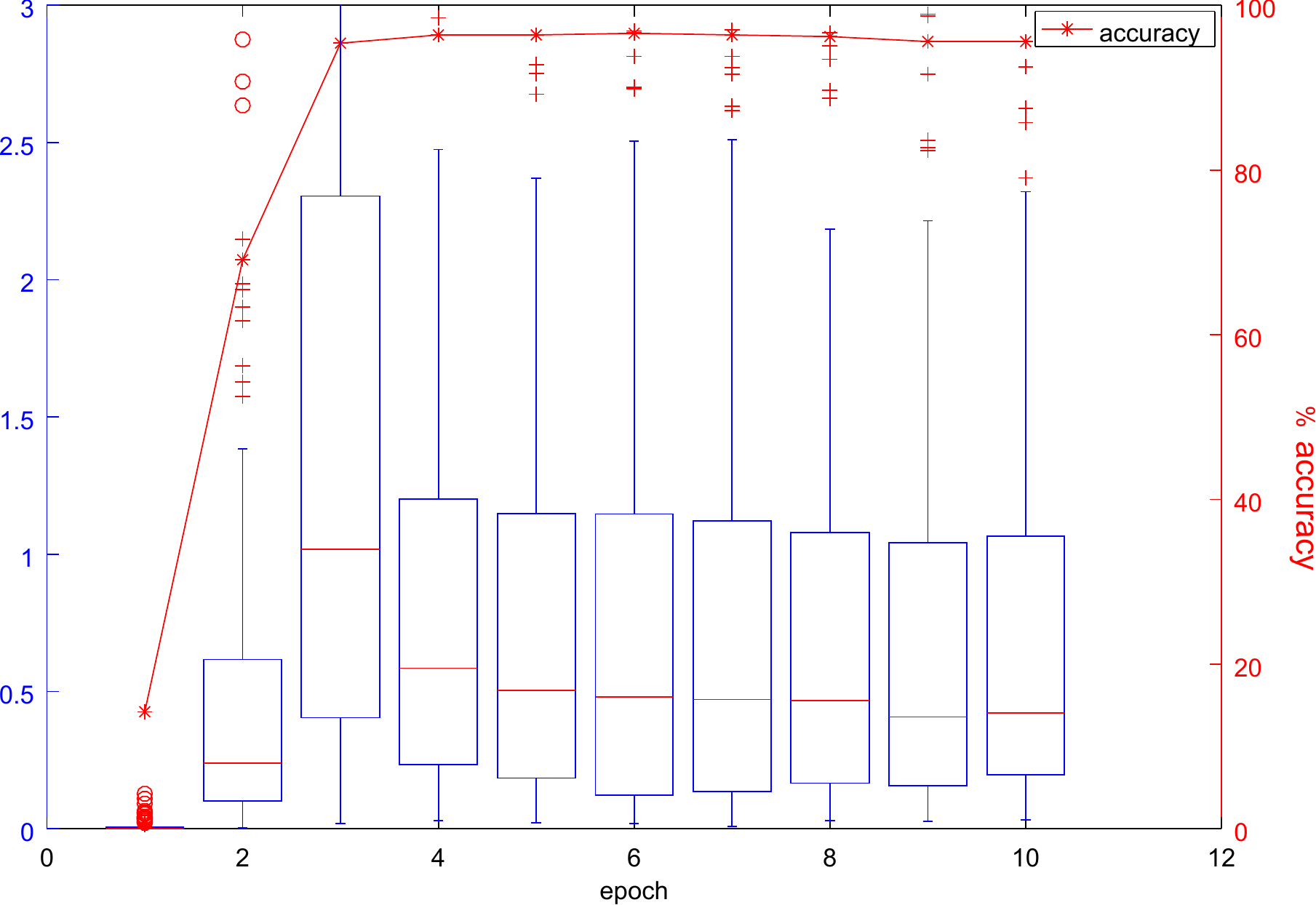}
                \label{fig lstm.lstm depth10}
        }
        \caption{Ratios at depth 10 in each epoch of training the RNN (a) 
        and the RLSTM (b). }
        \label{fig lstm.depth10}
\end{figure*}

We now examine whether the two models can encounter the vanishing gradient problem. 
To do so, we looked at the the back-propagation phase of each model in Experiment 1 on 
the third dataset (the one containing sentences of lengths from 21 to 30 tokens). 
For each tree, we calculated the ratio
\begin{equation*}
    \frac{\lVert \frac{\partial J}{\partial \mathbf{x}_{keyword}} \rVert}{\lVert \frac{\partial J}{\partial \mathbf{x}_{root}} \rVert}
\end{equation*}
where the numerator is the norm of the error vector at the keyword node and 
the denominator is the norm of the error vector at the root node. This ratio gives us an 
intuition how the error signals develop when propagating backward to leaf nodes:
if the ratio $\ll 1$, the vanishing gradient problem occurs; else if the ratio $\gg 1$, 
we observe the exploding gradient problem.

Figure~\ref{fig lstm.rnn backprop} reports the ratios w.r.t. 
the keyword node depth in each epoch of training the RNN model. 
The ratios in the first epoch are always very small. 
In each following epoch, the RNN model
successfully lifts up the ratios steadily (see Figure~\ref{fig lstm.rnn depth10}
for a clear picture at the keyword depth 10), but a clear decrease when the depth becomes larger 
is observable. For the RLSTM model (see Figure~\ref{fig lstm.lstm backprop}
and \ref{fig lstm.lstm depth10}), the story is 
somewhat different. The ratios go up after two epochs so rapidly that there are even some 
exploding error signals sent back to leaf nodes. They subsequently go 
down and remain stable with substantially less exploding error signals. 
This is, interestingly, concurrent with 
the performance of the RLSTM model on the development set 
(see Figure~\ref{fig lstm.lstm depth10}). 
It seems that the RLSTM model, after one epoch, quickly locates the keyword node in 
a tree and relates it to the root by building a strong bond between them via error signals. 
After the correlation between the keyword and the label at the root is found, 
it tries to stabilize the training by 
reducing the error signals sent back to the keyword node. 
Comparing the two models by aligning Figure~\ref{fig lstm.rnn backprop} with 
Figure~\ref{fig lstm.lstm backprop}, and Figure~\ref{fig lstm.rnn depth10}
with Figure~\ref{fig lstm.lstm depth10}, we can see that the RLSTM model is more capable of 
transmitting error signals to leaf nodes. 

It is worth noting that we do see the vanishing gradient problem happening 
when training the RNN model in Figure~\ref{fig lstm.rnn backprop}; but 
Figure~\ref{fig lstm.rnn depth10} suggests that the problem can become less 
serious after a long enough training time.
This might be because depth 10 is still manageable
for the RNN model. 
(Notice that in the Stanford Sentiment Treebank, 
more than three quarters of leaf nodes are at depths less
than 10.)
The fact the the RNN model still doesnot perform better than 
random guessing can be explained using the arguments given by 
\newcite{bengio1994learning}, who show that there is a trade-off 
between avoiding the vanishing gradient problem and capturing long
term dependencies when training traditional recurrent networks.

\section{Conclusions}
\label{section conclusion}

The experimental results show that the RLSTM is superior to the RNN in terms of  
overcoming the vanishing gradient problem and capturing long term dependencies. 
This is in parallel with general conclusions about the power of the LSTM architecture 
compared to traditional Recurrent neural networks. In future work we focus on more complex cases involving negation, composition, etc.

\bibliography{ref}
\bibliographystyle{acl2016}

\end{document}